\documentclass{nle}

\title[AQG Based on Sentence Structure Analysis using Machine Learning]
      {Automatic Question Generation Based on Sentence Structure Analysis using Machine Learning Approach}

\author[Miroslav Bl\v{s}t\'{a}k and Viera Rozinajov\'{a}]
{
   M.\ns B\ls L\ls \v{S}\ls T\ls \'{A}\ls K\ns  and  V.\ns R\ls O\ls Z\ls I\ls N\ls A\ls J\ls O\ls V\ls \'{A} \\
	Kempelen Institute of Intelligent Technologies\\
	Mlynske nivy~5, Bratislava, Slovakia
}

\received{21 Jun 2021 }

\pagerange{\pageref{firstpage}--\pageref{lastpage}}
\pubyear{2021}

\usepackage{graphicx}
\usepackage{algorithm}
\usepackage{algorithmic}
\usepackage{multirow}
\usepackage{url}
\usepackage{amsmath}

\begin{document}

\label{firstpage}
\maketitle

\begin{abstract}
Automatic question generation is one of the most challenging tasks of Natural Language Processing. It requires “bidirectional” language processing: firstly, the system has to understand the input text (Natural Language Understanding) and it then has to generate questions also in the form of text (Natural Language Generation). In this article, we introduce our framework for generating the factual questions from unstructured text in the English language. It uses a combination of traditional linguistic approaches based on sentence patterns with several machine learning methods. We firstly obtain lexical, syntactic and semantic information from an input text and we then construct a hierarchical set of patterns for each sentence. The set of features is extracted from the patterns and it is then used for automated learning of new transformation rules. Our learning process is totally data-driven because the transformation rules are obtained from a set of initial sentence-question pairs. The advantages of this approach lie in a simple expansion of new transformation rules which allows us to generate various types of questions and also in the continuous improvement of the system by reinforcement learning. The framework also includes a question evaluation module which estimates the quality of generated questions. It serves as a filter for selecting the best questions and eliminating incorrect ones or duplicates. We have performed several experiments to evaluate the correctness of generated questions and we have also compared our system with several state-of-the-art systems. Our results indicate that the quality of generated questions outperforms the state-of-the-art systems and our questions are also comparable to questions created by humans. We have also created and published an interface with all created datasets and evaluated questions, so it is possible to follow up on our work.
\end{abstract}

\section{Introduction}

Education is moving into the online environment and e-learning in combination with artificial intelligence is a very acute topic today \cite{ashraf2017,colchester2017}. From the educational point of view, knowledge evaluation is still a problematic part of e-learning. This is because human experts (e.g. teachers or lecturers) are currently irreplaceable during the knowledge verification phase. There are many systems for supporting education, but the task of question generation is weakly covered. Besides that, the utilization of generated questions goes far beyond education and learning. Question generation systems can also be used for the improvement of question answering systems \cite{reddy2017,tang2017} creating virtual agents for human-computer interaction and many other applications. So, the purpose of automatic question generation (AQG) systems is not limited to the area of education.

There are various approaches to the task of AQG. We are focusing on pattern-matching because thanks to the continuous development of Natural Language Processing (NLP) tools, this way offers a wide perspective (tools like part-of-speech taggers, named entity recognizers, semantic role labelers, etc.). Approaches based on pattern-matching leverage the patterns extracted from the text and use them for sentence-to-question transformations. We primarily use combinations of lexical patterns consisting of token sequences, syntactic patterns consisting of part-of-speech (POS) tags and semantic patterns consisting of semantic information about words (e.g. named entity tags or hypernyms). As manual creation of patterns for sentences is difficult and time consuming, we included machine learning techniques which allow patterns to be obtained automatically from an initial set of sentence-question pairs. This concept is also known as the data-driven approach (learning from data). The main contributions can be summarized as follows: 

\begin{enumerate}
  \item we propose a novel method for the task of question generation from text which uses a combination of the traditional linguistic approach known as pattern matching with actual machine learning approaches;
  \item we present a data-driven approach in which transformation patterns are automatically extracted from samples of sentence-question pairs (this approach allows the system to continuously learn to create new questions for new unseen sentences);
  \item we suggest storing extracted patterns in a hierarchy of patterns which allows us to find the most common type of question for a specific sentence (this approach should also be beneficial in related NLP tasks).
\end{enumerate}

The rest of this paper is structured as follows: 
\begin{itemize}
  \item In the second section, we introduce contemporary approaches used in question generation systems, our proposed categorization of state-of-the-art systems based on several criteria and an evaluation of generated questions in these systems with a focus on the domain of factual questions which are generated from unstructured text. 
  \item In the third section, we describe our proposed framework for generating questions from unstructured text including a detailed description of all steps of the process from preparation of a dataset to question generation and its evaluation thorough comparisons. 
  \item In the fourth section, we show and explain our experiments and discuss the evaluation of results from various points of view. 
  \item In the final section, we make our conclusion and propose suggestions for future work.
\end{itemize}

%
%
%
%

\section{Related works}
It is difficult to clearly categorize related works because there are several criteria and points of views to consider. If we focus on education, one of the most important criteria for differentiation is the presence of the correct answer. Many question generation systems generate questions regardless of whether the answers are present in the input text or not. And there are also systems generating questions which can be answered without understanding the text (e.g. they utilize the knowledge of the language grammar \cite{chen2006,mostow2012,lin2007,kunichika2011}). We primarily focus on factual questions. Factual questions are based on the facts which are mentioned in the text. 

\subsection{Factual Question Generation}
Considering the extent of input text, we can classify question generation systems into those that create questions from structured data \cite{daquin2011,olney2012,serban2016}, a combination of structured and unstructured data \cite{yao2012,lin2009} or from a specific part of the unstructured data. The specific part can be: 
\begin{itemize}
	\item a sequence of a few words \cite{agarwal2011,becker2012,brown2005,karamanis2006,kumar2015}, 
	\item a sentence \cite{afzal2014,agarwal2011b,mazidi2016,curto2012,ali2010,heilman2010,mitkov2006,chali2018},
	\item a paragraph \cite{mannem2010}, Lin et al. (2009),
	\item a full article or topic \cite{yuan2017,du2017,jin2016,chali2012,chali2015}.
\end{itemize}

One of the most typical approaches for generating questions based on a few words is known as fill-in-the-blank (Afzal and Mitkov (2014), Lin et al. (2007), Kumar et al. (2015)). These systems first identify some part of the text (one word or a sequence of several words) which will be an answer and remove it from the text. The words to be removed are usually chosen by part-of-speech (POS) tags (e.g. adjectives, prepositions or nouns). Identification of the gap is the most challenging part of this process and generation of set of distractors is also very complicated. 

Questions generated from one sentence or from a sequence of sentences require the engagement of more sophisticated linguistic techniques. There are systems which generate questions as interrogative sentences transformed from input declarative sentences. This process is also known as sentence-to-question transformation and it usually uses a set of transformation rules. There are several approaches used in related research papers.

One of the first factual question generation system based on sentence-to-question transformation was introduced in Mitkov \shortcite{mitkov2006}. They focus on a methodology to generate multiple-choice test items about facts explicitly stated in single declarative sentences. In Heilman \shortcite{heilman2010}, the authors extended this idea and used syntactic patterns created from parse tree representation. They firstly extract simplified statements from a text and then transform the declarative simplified sentence into a question. Sometimes, they utilize semantic information about words to decide if the noun reflects a person or a thing (if it is possible to use the interrogative pronoun who or what). An approach based on modifying the parse tree was also used in \cite{kalady2010} with a combination of keywords extraction. In addition, one system in the QGSTEC challenge uses the parse tree to generate questions \cite{yao2010} (tree-based pattern matching rules).

There are several approaches which leverage basic syntactic and semantic sentence patterns. In Ali \shortcite{ali2010}, their obtained sentences were matched to a predefined group of syntactic patterns and a set of possible questions was assigned to each pattern. A combination of lexical and syntactic patterns was used in Curto \shortcite{curto2012}. They created an interactive virtual agent for generating questions about artifacts in a museum. It learned patterns from a set of sentence-question pairs and they were subsequently evaluated by Google search. A strong limitation of this work is that questions are focused only on the domain of art and the coverage of sentences from the dataset was weak. Several systems based on rules which combine semantic role labeling and named entities were presented in other research papers (e.g. Pal et al. 2010 or \cite{lindberg2013}). In our previous work \cite{blstak2017}, we introduced an approach combining lexical, syntactic and semantic patterns where we also showed the possibilities of using these patterns together. According to our knowledge, except for our previous work, there is no system based on transformation patterns which use the data-driven approach for extraction of rules during the learning phase (this does not apply to systems using the neural approach as we will mention later). Another question generation framework (QGaSP), which leverages different levels of linguistic information was introduced in \cite{rodrigues2016}. They used a similar approach to the virtual agent about artifacts, but they also used semantic role labeling. In \cite{das2016}, the authors focus on the problem of increasing the number of acceptable questions when sentences are too complex, or patterns are ambiguous. They created a set of rules which split sentences into segments based on syntactic patterns, in order to assist the generation of the correct type of questions. Another improvement to pattern-matching processes in AQG was suggested by \cite{mazidi2016}. Here, an approach was presented combining extracted syntactic information from sentences with semantic role labelling and syntactic dependency parsing. In \cite{song2016}, the authors tried to automate the process of template creation with the intent of reducing human effort. They manually created a small number of hand-crafted templates in the form of predicates with placeholders for the subject and object for question generation from knowledge base. These seed questions were then expanded through search engines (e.g. Google or Bing) as a query. It is very difficult to determine the benefit of this approach due to weak experiments and comparisons. Evaluation was done only on a question generation dataset consisting of triplets from Freebase even though this dataset was generated by machine, so it is obvious that the questions are unnatural as authors have mentioned. They also experimented with a dataset designed for the classification of short documents, but this is a different task to question generation and they only checked if questions were natural by using results obtained from human evaluators. The authors also did not consider if the generated questions were answerable. Even though the seed set of patterns must be created by humans, it is an interesting way of introducing automation into the process to reduce user fatigue.

\subsection{From linguistic approaches to statistical approaches}
When we look at NLP tasks globally, there is one important aspect regarding the way of approaching the given problem. We distinguish between linguistic approaches and statistical approaches (often assigned to machine learning methods). We consider today's popular approaches based on artificial neural networks as a subcategory of statistical approaches whereas the neural network model also takes into account statistical occurrences from the training dataset. In several NLP tasks, machine learning approaches beat linguistic ones (e.g. machine translation task), but this requires large datasets for training and evaluating the system. 

Obtaining datasets for AQG tasks is not particularly simple. There is one large question dataset called 30m factoid question-answer corpus in Serban et al. \shortcite{serban2016}. It is used in the question generation task, but it contains 30 million questions mapped to triplets, not to sentences. However, the IBM Watson system uses this dataset in the question answering task. According to their first experiments, using this auto-generated dataset was effective for training \cite{lee2016}. 

For the task of question generation from unstructured text, there is only one small dataset called QGSTEC2010 \cite{rus2012}. It consists of 90 sentence-question pairs for training and another 90 pairs for the testing phase. Evaluation on this dataset will be described in the next section. However, a dataset which consists of just 90 questions is not suitable for use for training models, and several researchers have leveraged self-made datasets from Wikipedia articles or datasets from related tasks (e.g. question answering).

Approaches using neural networks began to become popular in question generation tasks with the publication of the dataset for the question answering task known as SQuAD - The Stanford Question Answering Dataset \cite{rajpurkar2016}. It consists of more than 100,000 questions from more than 500 articles. Although it was not originally intended for the question generation task, it was adopted for it in several recent works \cite{du2017,song2018,wang2018,zhou2018}. This transition to a robust dataset brings new problems, especially in the area of evaluation. Firstly, it is necessary to split the dataset into train and test subsets with similar characteristics in each (e.g. distribution of question types) and then to establish how to simplify the process of evaluation because it is no small task for a human to evaluate so much generated data. Another possible resource of data for this task is project LearningQ \cite{ICWSM18LearningQ}\footnote{available at \url{https://github.com/AngusGLChen/LearningQ}}. It is a dataset which can be used for educational question generation whereas it contains about 7000 questions from TED-Ed and more than 200000 questions from Khan Academy.

In Du et al. \shortcite{du2017}, an approach for generating questions from text passages with the intent of revealing reading comprehension was proposed. The authors used an attention-based sequence learning model and compared two versions of encoder-decoder architecture: in the first, they encoded only one sentence and in the second, they also added paragraph-level information. Both versions used the bidirectional LSTM network. The authors conducted experiments on a modified SQuAD dataset as follows: they extracted sentences and paired them manually to the questions and then pre-processed the text with the Stanford CoreNLP tool. The dataset was randomly divided into a training set, a development set and a test set (80/10/10). Human evaluation was conducted on 100 randomly selected sentence-questions pairs. The model which only encodes sentence-level information achieved the best performance across all metrics (BLEU-3, BLEU-4 and METEOR).

A similar approach was used in Zhou et al. \shortcite{zhou2018}. Here, the authors also proposed a neural encoder-decoder model for generating questions from natural language sentences and conducted experiments with the SQuAD dataset. They used bidirectional GRU to build an encoder and an attention-based decoder. They did not only used sequences of words, they also leveraged handcrafted features like lexical features (word cases, POS tags and NER tags) and an answer position indicator. They extracted sentence-question-answer triplets from the SQuAD dataset and compared several versions of neural networks. Their results showed that the model with a configuration consisting of a pre-trained word embedding matrix provided the best results. In human evaluation, the authors used BLEU metrics and users were asked to evaluate the questions generated from the best model. In addition, they evaluated the benefits of selected features and the results showed that answer position indicator plays a crucial role in generating correct questions. Other features (word case, POS and NER) also contribute to this process. 

In Song et al. \shortcite{song2018}, the authors tried to determine if context information can improve the question generation task. Their encoder-decoder framework also uses bidirectional LSTM - the encoder reads the passage and answer and the decoder generates a set of questions. They compared three strategies to match the passage with the answer. In full-matching, the authors encoded all words and the word order, in attentive-matching they considered the weighted sum of all answers and all words without word order, and in the max-attentive-matching configuration, they considered only the most relevant answer state. The authors compared their model with a basic sequence-to-sequence model and their experiments showed that a combination of all the strategies performed best. This is because these strategies are complementary. The best strategy is the full-matching strategy. In \cite{sun2018}, the authors tried to handle issues where the answer type is not consistent with the question type. They introduced two models to deal with this problem: an answer-focused model which takes an answer as additional information and a position-aware model which uses distance between the context words and the answer. In \cite{kumar2018}, a framework for automatic question generation from text using deep reinforcement learning was presented. During the training phase, the authors have used evaluation metric as a reward (e.g. BLEU or ROUGE-L) and their approach outperforms previously mentioned works (the comparison will be discussed in the next section). Based on the findings of these works focused on neural networks, we can say that question generation approaches built on seq2seq models work better with additional lexical features (POS, NER, answer position or pre-processed word embedding matrices). In these works, these were used for generating questions from passages of text instead of standalone sentences. However, there are some crucial disadvantages resulting from the use of neural networks. These models are hard to interpret, so it is not easy to improve by adding new sentence-question pairs as in pattern matching approaches or machine learning being improved by instant feedback after deployment (e.g. rewards in reinforcement learning or new sentence-question pairs obtained later). And sometimes, models have to be re-trained when adding new training data.

\subsection{Evaluation of the Questions}
Evaluation of AQG systems is not an easy task - and sometimes it can be particularly tricky. As it was found in the recent review of the evaluation methodologies used in AQG based on 37 papers \cite{amidei2018}, there is still no standard way of evaluating AQG systems. Generally, there are two possibilities for how to evaluate systems and generated questions. The first approach is to obtain several articles, generate questions from these articles and quantify the total number of correctly-generated questions with metrics like precision, recall and F1 measure. This type of evaluation dominates (e.g.: (Kunichika et al. 2011), \cite{heilman2010} and many others). The sources of text could be:

\begin{itemize}
	\item school textbooks \cite{susanti2015,kumar2015} or (Kunichika et al. 2011),
	\item online encyclopedias such as Wikipedia or Britannica \cite{heilman2010,agarwal2011b,chali2015},
	\item community question-answering systems e.g. Stack Overflow or Yahoo Answers (Rus et al. 2012),
	\item question-answering datasets TREC2007 or SQuAD.
\end{itemize}

The second option is to evaluate systems utilizing a dataset created for this task which is then used across various systems. As we know, there is only one dataset which keeps tuples of sentences and questions for this purpose. It was first used by participants in the First Question Generation Shared Task Evaluation Challenge (QGSTEC) by participants of this event \cite{pal2010,varga2010,yao2010} and Ali et al. (2010) and later in several research papers \cite{chali2015,agarwal2011b} or in Curto et al. (2012). The evaluation results are summarized in table \ref{tablecomparison1}. The first four systems were evaluated directly at QGSTEC  by human evaluators. We simply transformed points score into a percentage and the number of the correct questions was calculated as the number of generated questions multiplied by correctness. In the case of (Curto et al. 2012), the focus was on questions which were related only to a specific domain (history) and in \cite{agarwal2011b}, a system which generates questions using discourse connectives was presented. The maximum number of questions was 121 because these types of questions should be generated only from the subset of input sentences (only sentences with discourse connectives e.g. since, when, or although). So, we calculate the ratio as 63 generated questions out of 121 possible questions. 

\begin{table}
  \caption{Comparison of AQG approaches on QGSTEC dataset}
  \begin{minipage}{\textwidth}
    \begin{tabular}{lcccc}
    \hline\hline
	System &
	Correctness (\%) &
	Q/sentence &
	Correct q. &
	Generated q. \\
    \hline
    Participant 1\footnote{\cite{yao2010}} 		&0\hpt64 &4\hpt37 &227 &354 \\
    Participant 2\footnote{\cite{varga2010}} 	&0\hpt34 &2\hpt04 &56 &165 \\
    Participant 3\footnote{\cite{pal2010}} 		&0\hpt30 &2\hpt58 &63 &209 \\
    Participant 4\footnote{(Ali et al. 2010)} 		&0\hpt21 &2\hpt07 &35 &168 \\
    Curto et al.\footnote{(Curto et al. 2012)} 		&0\hpt56 &0\hpt31 &14 &25 \\
    Agarwal et al.\footnote{\cite{agarwal2011b}} 	&0\hpt7  &0\hpt52 &44 &63 \\
    \hline\hline \\
    \end{tabular}
    \vspace{-2\baselineskip}
  \end{minipage}
  \label{tablecomparison1}
\end{table}

There is one other option when comparing generated questions with questions created by users. We can use text similarity metrics as presented in (Du et al. 2017) or \cite{divate2017}. In several related Natural Language Generation tasks (such as machine translation), there are widely-used metrics for the evaluation of text generated by machines to text expected by users. For example, BLEU \cite{papineni2002}, METEOR \cite{denkowski2011} and ROUGE \cite{lin2004} are very popular but, for this comparison, we also need a referential set of questions obtained from human evaluators. For this purpose, a created user interface or paid evaluation services like Mechanical Turks are usually used \cite{buhrmester2011,mazidi2016}.

%
%
%
%
As we mentioned in the previous section, question generation approaches based on neural networks are usually evaluated on the basis of the SQuAD dataset (Du et al., 2017; Zhou et al., 2017; Song et al., 2018; Kumar et al. 2018). There are two possible types of evaluations: evaluation by humans and automatic evaluation. When the dataset is large, it is not possible for humans to evaluate all the questions, so human evaluation is usually done only in a subset of instances. In Kumar et al. \shortcite{kumar2018}, the authors made a comparison from both sides. Summarization of the most important results (from the best models) are shown in table \ref{tablecomparison2} (results obtained from the original papers). 

\begin{table}
  \caption{Automatic evaluation of AQG approaches on SQuAD dataset}
  \begin{minipage}{\textwidth}
    \begin{tabular}{lccccc}
    \hline\hline
	System &
	BLEU-1 &
	BLEU-2 &
	BLEU-3 &
	BLEU-4 &
	ROUGE-L \\
    \hline
    Du et al. \shortcite{du2017} 							&43\hpt09 	&25\hpt96 &17\hpt50 &12\hpt28 &39\hpt76 \\
    Song et al. \shortcite{song2018}							&- 			&- 			&- 			&13\hpt98 &42\hpt72 \\
    Zhou et al. \shortcite{zhou2018}							&- 			&- 			&- 			&13\hpt29 &- \\
    Kumar et al. \shortcite{kumar2018} \footnote{model with reward function based on BLEU}  		&46\hpt59 	&29\hpt68 &20\hpt79 &15\hpt04 &41\hpt73 \\
    Kumar et al. \shortcite{kumar2018} \footnote{model with reward function based on ROUGE-L}  		&48\hpt13 	&31\hpt15 &22\hpt01 &16\hpt48 &44\hpt07 \\
    Sun et al. \shortcite{sun2018}  \footnote{hybrid model}							&43\hpt02 	&28\hpt14 &20\hpt51 &15\hpt64 &- \\    \hline\hline \\
    \end{tabular}
    \vspace{-2\baselineskip}
  \end{minipage}
  \label{tablecomparison2}
\end{table}

In Zhou et al. \shortcite{zhou2018}, the authors used a three-level scale from 1 (bad questions) to 3 (good questions). Three human raters labeled 200 questions sampled from the test set and obtained an average score of 2.18 points in comparison to the baseline of 1.42. In Kumar et al. \shortcite{kumar2018}, the authors used a percentage scale (0-100) and evaluated three criteria: syntax, semantics and question relevance. Three human raters labeled 100 randomly selected questions. The best scores for chosen criteria were 84 (syntax), 81.3 (semantics) and 78.33 (relevance). In Du et al. \shortcite{du2017}, human evaluators used a scale of 1-5 and their criteria were naturalness (grammaticality and fluency) with a score of 3.91 and difficulty (sentence-question syntactic divergence and the reasoning needed to answer the question) with a score of 2.63. In \cite{hosking2019}, three human evaluators rated 300 generated questions on a scale of 1-5 and criteria were fluency of language and relevance of questions. 

There is another way to automatically evaluate generated questions – extrinsic evaluation. It is used to detect how useful the generated questions in real application are. As the purpose of factual questions is in education and validation of knowledge, it is appropriate to verify their contribution in this field. For example in \cite{skalban2013}, the extrinsic evaluation (described in 5.2.) aims at evaluating usefulness of generated questions in real life applications and they evaluate generated questions also with focus to these attributes:
\begin{itemize}
\item Whether the presence of text-based pre-questions helps test-takers to answer post-questions more accurately (i.e. more questions are answered correctly).
\item Whether the presence of pre-questions with screenshots extracted from the video helps the test-takers to answer post-questions more accurately.
\item Whether the presence of text-based pre-questions affects the time taken to answer post-questions.
\item Whether the presence of pre-questions with screenshots extracted from the video affects the time taken to answer post-questions.
\end{itemize}

The most common model for question classification in terms of fulfilling some educational goals is probably Bloom taxonomy, which was invented in 1959 and revised in 2001 \cite{bloom1956,bloom2001}. This taxonomy defines six levels of questions based on the complexity of created questions. However, utilizing sentence-to-question transformation for generating these questions is feasible only for the first and the second level (in most cases only for the first one). Simply put, the questions of the first level (remembering) are questions based on repeating the facts from the text. The questions of the second level (understanding) are questions based on information that is explicitly present in the text, but they may be interpreted by different words (e.g. synonyms, hypernyms etc.). In case of generating questions from the text by transforming declarative sentences into questions, it seems better to use the three-level classification for reading comprehension area proposed by \cite{pearson1978}. The first level of questions are those, which are explicitly mentioned in the text, the second level of questions can be answered by combining several facts from the text and the third level of questions involve some background knowledge. There are some other useful taxonomies for question generation tasks. In \cite{Day2005DevelopingRC}, the authors focus on developing reading comprehension questions and some attributes can be also used as a checklist for teachers. In \cite{graesser1992}, several question-generation mechanisms were proposed. They categorized questions (inquiries) on the basis of semantic, conceptual and pragmatic criteria into 18 categories. The most common questions generated by AQG systems belong to categories: verifications (Is the fact true?) and concept completion (Who?, What?, etc.). 

\subsection{Summary of Related Works}
We have shown that there are various approaches and various AQG systems in areas of computer-aided learning and question generation. We have mentioned the most interesting among them which are related to our work. As we are focusing on factual question generation from unstructured text, we firstly tried to make an overall categorization of state-of-the-art approaches in the area of AQG from several points of view and we then identified and determined the sub-areas of our direction. There are several related systems we can compare with.
We also mention several papers focused on neural networks. These also perform better with additional lexical features such as a list of POS labels, entities, answer position or pre-processed word embedding matrices. However, these models are hard to interpret and difficult to improve by feedback information as they stand as a black box. In pattern-matching approaches, we can improve a model simply by adding new sentence-question pairs or marking some question as incorrect so it will not be used in the future. 

We have also mentioned some existing datasets and sources of texts which are often used in the question evaluation phase. There are three possible options for how to evaluate AQG systems. Firstly, we can calculate the correctness of generated questions from a specific type of text. Then, for mutual comparisons of various systems, we can use question generation datasets and compare the questions generated from them. And, finally, if we also want to compare the quality of questions from a user perspective, we can evaluate the similarity of questions generated by systems with questions expected by users.

In the next section, we will describe our framework for factual question generation and compare it with state-of-the-art-systems based on these criteria.


\section{Our Proposed Question Generation Approach}
Our main contribution to AQG consists of the data-driven approach and a combination of traditional linguistic approaches used in NLP tasks with machine learning approaches used during the system-training and system-improvement steps. When we look at similar areas, it is the machine translation task which has lately gained the greatest attention. The quality of translated text increased rapidly after traditional linguistic methods were supplemented (and/or replaced) by machine learning approaches. Thanks to the availability of large corpora of multilingual parallel texts, the learning process of translation is more efficient. In our task, we adopted this data-driven approach for training the system to generate questions. 

\begin{figure}
  	\vspace{0.2cm}
	\caption{Conceptual overview of our proposed AQG framework}
	\includegraphics[totalheight=5.3cm]{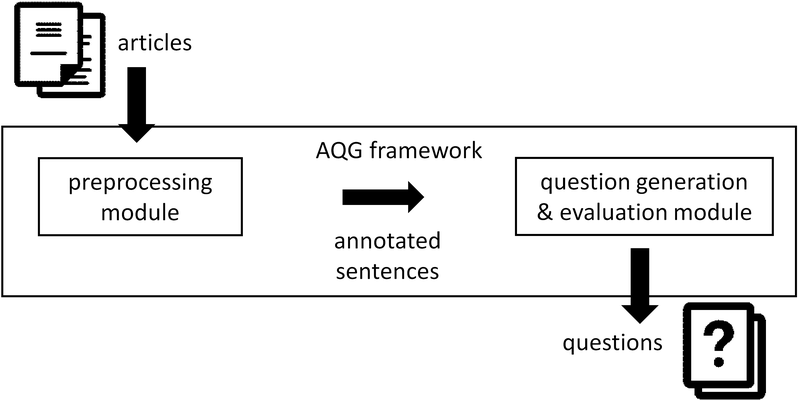}
	\label{figure_aqgfw}
\end{figure}

Our proposed framework for AQG (figure \ref{figure_aqgfw}) uses datasets consisting of sentence-question pairs extracted from text. Firstly, preprocessing extracts features from sentences. Then we create a special data structure which keeps these features and calculates the possible sentence transformation rules for transformation of a sentence into a question. When questions are created, the question estimator module estimates the quality of the generated questions. We adopted the reinforcement learning approach to improve our system by reflecting explicit feedback about the quality of the questions generated in the past. In the next sections, we describe these steps in detail.

\subsection{Sentence Preprocessing}
Initial preprocessing steps are required because we need to represent input of unstructured text as the set of features in order to use it for machine learning. After sentence boundary identification and token extraction, we obtain several pieces of information about each token from the sentences. These data are stored in the data structure that we named ``composite pattern'' (CP). This is the structure of a sentence represented as sequence of tokens and its labels (syntactic and semantic). We can imagine the composite pattern for one sentence as a matrix where rows are individual patterns of sentences and each token should have labels assigned to this pattern based on an annotation tool (the columns of the matrix). The structure of composite pattern is illustrated in table~\ref{table1}. 

We experimented with various sentence patterns consisting of various information about tokens. After a few experiments, we found the most important features for this task are:
\begin{itemize}
  \item lemma: sequence of lemmas (significance of lemma is in cases of words from closed word categories like prepositions, articles or pronouns),
  \item POS: sequence of part-of-speech tags,
  \item POS simple: simplified POS tags (tags grouped into general categories, which is more effective for mapping in the case of various sentence structures),
  \item NER: information about named entities,
  \item GKG: semantic information obtained from Google Knowledge Graph service,
  \item Viaf: search for entity categories in Virtual International Authority File,
  \item SST: semantic tags from Wordnet lexicon.
\end{itemize}
\begin{table}
  \caption{Example of Data Structure for Composite Pattern (CP), which consists of several patterns (rows of table) }
  \begin{minipage}{\textwidth}
    \begin{tabular}{lccccc}
    \hline\hline
	\bf{Sentence}&
	\bf{Peter}&
	\bf{Sagan}&
	\bf{comes}&
	\bf{from}&
	\bf{Slovakia} \\
    \hline
    \bf{Lemma} 		& Peter	&Sagan	&come	&from	&Slovakia 	\\
    \bf{POS} 		& NNP	&NNP	&VBZ		&IN		&NNP	\\
    \bf{POS Simple} 	& NN 	&NN 		&VB 		&IN 		&NN 		\\
    \bf{NER}		& person	&person 	& - 		& - 		&location 	\\
    \bf{GKG} 		& person	&person 	& - 		& - 		&country 	\\
    ... 			& ...	&...	&...	&...	&...	\\
    \hline\hline \\
    \end{tabular}
    \vspace{-2\baselineskip}
  \end{minipage}
  \label{table1}
\end{table}
We conducted several experiments to explore the importance of these features. Firstly, the sequence of POS labels keeps information about sentence structure and allows us to recognize the syntactic structure of sentence. The positive influence is obvious and it is also used in many approaches leveraging transformation rules (e.g. Curto et al. \shortcite{curto2012}, Heilman et al. \shortcite{heilman2010}, Ali et al. \shortcite{ali2010} and many others). After a deeper analysis of sentence classification, we added a general version of a POS pattern (marked as POS simple). We wanted to group several POS patterns into one because some labels are mutually substitutable for the question generation task. For example, there is no big difference in singular or plural noun forms, or between past and present verb tenses. Question generation for both cases is similar. But the difference between verb and noun is crucial. We simply replaced substitutable labels into one single label and called it POS simple pattern (e.g. all verb forms are merged into the verb instead of verb in past tense, verb in present tense, etc. labels). The first level of hierarchy of patterns was created in this way. There are one-to-many relations between POS simple and POS patterns. This hierarchical grouping also helps us to search for related patterns if the sentence pattern is not yet known. 

We also analyzed NER, GKG, Viaf and SST labels separately and found NER and GKG to be the most reliable. Their evaluation also depends on dataset. Therefore, we used them together and in cases where they do not match and stored both outputs, as we will describe later. While SST labels bring minor improvement, Viaf labels bring benefits in cases where some concepts are mentioned in different or original language (e.g. the name of a non-English person or the name of a city in the original language). It is also applicable for deduplication, which is useful in question generation from sentences. We tried to find research papers focused on comparing the accuracy of individual tools, but  found only papers focused to knowledge graphs (e.g. \cite{gawriljuk2016} where they compared several ontology sources). However, we mainly relied on our findings obtained empirically by side experiments for our previous papers.

As working with multiple patterns is computationally expensive, we suggested a hierarchy of patterns in our previous research \cite{blstak2016}. We have shown that this approach could be beneficial not only for efficient pattern matching by traversing in the hierarchy, but it also overcomes the problem of weak pattern coverage. This is due to the fact that we can more easily find patterns with similar structure (more general pattern). We have also demonstrated it on a combination of several basic sentence patterns: POS pattern (a sequence of POS labels), NER pattern (a sequence of named entities), SST pattern (a sequence of super-sense tags obtained from Wordnet) and Linked Data pattern (which consists of concepts found in Freebase). In this work, we made several changes. Firstly, we added a new top-level pattern and removed/merged patterns with low impact. We extended the pattern hierarchy with the top-level pattern called POS simple. This allowed us to increase coverage of sentences as we have mentioned before. We added a GKG pattern (concepts obtained from Google Knowledge Graph) instead of a Linked Data pattern (the Freebase project was cancelled, and its data are now used in GKG). Although GKG and NER are similar, we used them in combination with one another: if the label is different, we decreased the estimated value of correctness for a question which relies on this information (for example: when we generate a question for a person and not all of the patterns  identify the entity of the person type in sentence).

Some patterns may have hierarchical relationships among each other (figure \ref{figure1}). For example, the POS simple pattern should be the parent node for a group of POS patterns. This means that it covers various POS patterns, so there is a set of sentences which may have the same POS simple' pattern and several different POS patterns. This principle also works for the relationship between POS and NER patterns where entities are almost exclusively nouns. For example, these sentences are different only in NER patterns and the remaining patterns are exactly the same (labels of sentence tokens from the left are: determiner, adjective, preposition, noun, verb, noun: 

\begin{figure}
  	\vspace{0.2cm}
	\caption{Hierarchical relationships between patterns: one POS simple pattern covers several various sentence subpatterns with higher level of granularity. }
	\includegraphics[totalheight=4.6cm]{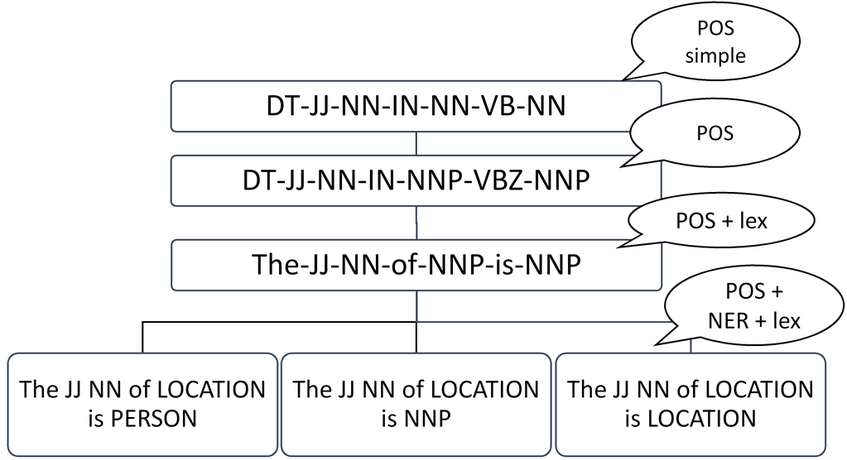}
	\label{figure1}
\end{figure}
  
\begin{itemize}
\item The largest city in Austria is Vienna.
\item The main currency in Europe is the Euro.
\item The current president of Slovakia is Andrej Kiska. 
\end{itemize}
The last token of all sentences has a different NER label: 'Andrej Kiska' is a person, 'Euro' is a currency (which is not detected as a named entity) and 'Vienna' is a location. Figure \ref{figure1} shows the difference in a hierarchy diagram. The POS simple pattern is the most generic pattern and it covers various types of sentences. We created this pattern for one special purpose: quick filtering of patterns from an indexed list. If the sentence does not match this pattern, we can consider the complex pattern as not applicable for the sentence. On the other hand, sequence of lemmas is the most specific pattern. These traits allow us to search for related patterns. If the system could not find rules for a specific sentence, it could use a more generic pattern allowing it to create at least some more general question. The construction of a complex pattern for a sentence is shown in procedure \ref{alg1} and the process of text preprocessing and the assigning of complex patterns to the sentences is shown in figure \ref{figure_fw_preprocessing}.

\begin{algorithm} 
\floatname{algorithm}{Procedure}
\renewcommand{\algorithmicrequire}{\textbf{Input:}}
\renewcommand{\algorithmicensure}{\textbf{Output:}}
\vspace{0.1cm}
\caption{Creation of complex pattern  (createCP)}
\label{alg1} 
\begin{algorithmic}[1]
    \REQUIRE sentence (string)
    \ENSURE complex pattern of sentence (data structure)
\vspace{0.1cm}
    \STATE $CP \Leftarrow \O$
    \FORALL{pattern \textbf{in} \{POS, NER, SST ...\}}
    	\STATE{$labels \Leftarrow \O$}
    		\FORALL{token \textbf{in} sentence}
    		\STATE{$labels \Leftarrow labels+ (token, pattern.obtainLabelForToken(token))$} 	    
    	\ENDFOR
    	 \STATE $CP \Leftarrow CP + labels$
    \ENDFOR    
    \RETURN CP
\end{algorithmic}
\end{algorithm}

%
%
%
%

\subsection{Learning phase}
The proposed question generation model is based on two machine learning approaches: supervised learning and reinforcement learning. First, we trained the model on pairs of sentences and questions associated with these sentences. These pairs are used to learn how to transform declarative sentences into questions. Each learned transformation consists of a pair of composite patterns: a composite pattern for input sentence and a composite pattern for an associated question. Transformation rules (TR) keep information on how to modify sentences to achieve the form of a question. These use a basic modification operation with sentence tokens: removing token, adding token, moving token or changing the form of token. Changing the token form is a special operation used in some cases where the word form must be transformed (e.g.: transforming the plural form into singular or changing past tense to the present). Training the model is simple (procedure \ref{alg2}), we simply obtained composite patterns for sentences and questions and stored this tuple in a list of unique transformation rules.

\begin{figure}
  	\vspace{0.2cm}
	\caption{Conceptual view of preprocessing module which extract the sentence from input text and creates composite patterns for each sentence. In case of complex sentences, there is also sentence simplification process which splits the sentence and add new simplified sentences into global list of sentences. }
	\includegraphics[totalheight=10.59cm]{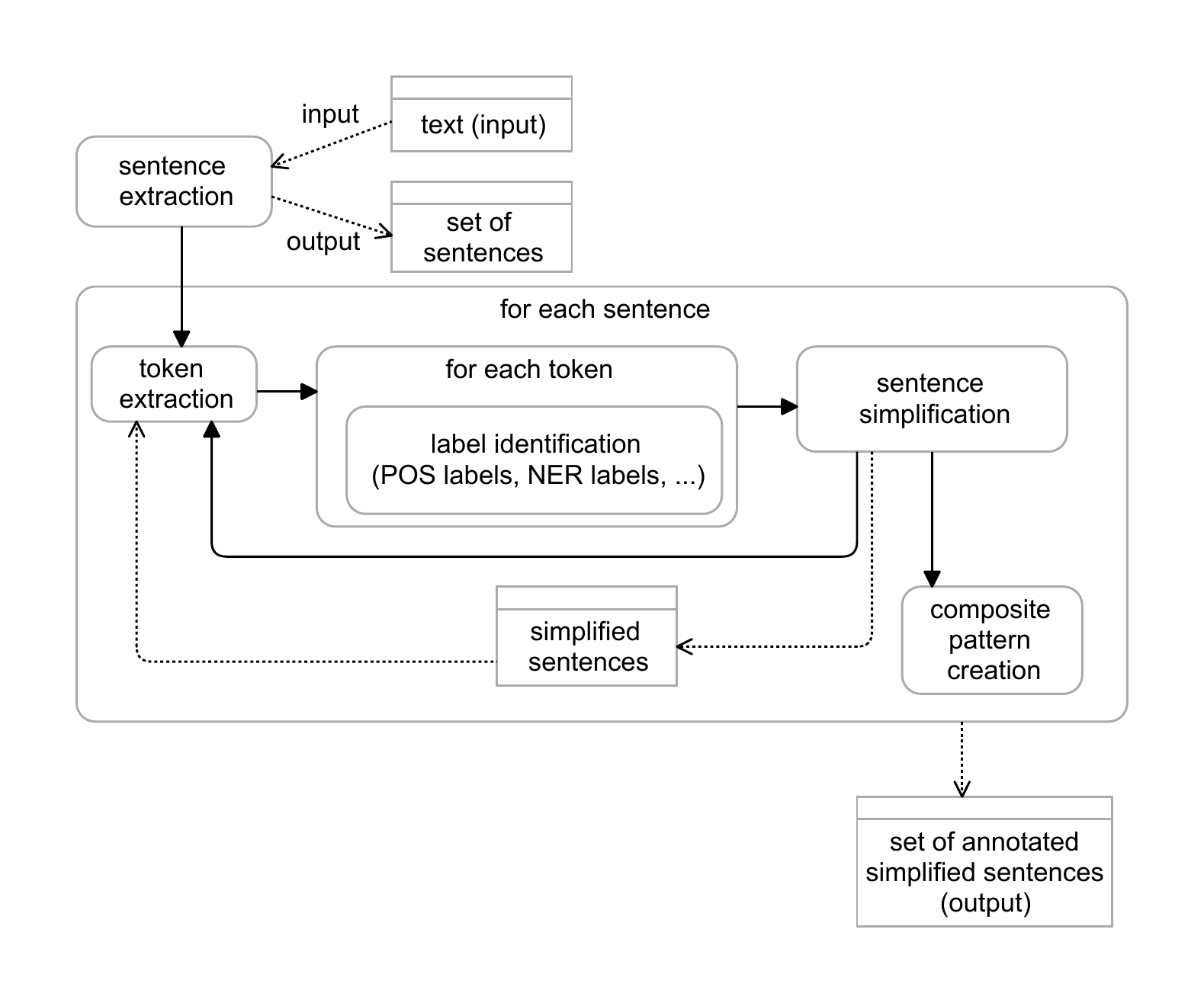}
	\label{figure_fw_preprocessing}
\end{figure}

\begin{algorithm} 
\floatname{algorithm}{Procedure}
\renewcommand{\algorithmicrequire}{\textbf{Input:}}
\renewcommand{\algorithmicensure}{\textbf{Output:}}
\vspace{0.1cm}
\caption{Training phase}
\label{alg2} 
\begin{algorithmic}[1]
    \REQUIRE dataset (list of sentence-question pairs)
    \ENSURE transformation rules
\vspace{0.1cm}
    \STATE $TR \Leftarrow \O$
    \FORALL{sentence,question \textbf{in} dataset}
    	\STATE{$cp\_sentence = createCP(sentence)$}
    	\STATE{$cp\_question = createCP(question)$}
	\IF{$(cp\_sentence, cp\_question)   ~ \NOT~\textbf{in}~~TR $} 
		\STATE {$TR = TR + \{(cp\_sentence, cp\_question)\}$}   	
	\ENDIF
    \ENDFOR    
    \RETURN TR
\end{algorithmic}
\end{algorithm}
\vspace{0.1cm}
The output of the initial step of the training phase is a list of tuples of composite patterns (a composite pattern for the sentence and a related composite pattern for the question). Their initial score value is set to 1 if the pattern is obtained from training data (because we consider the sentence-question pairs from training data as correct). Then, the score is updated periodically. This score is a product of four components (number between 0 and 1):
\begin{itemize}
\item similarity between CP of input sentence and CP of sentence in stored transformation rule,
\item similarity between CP of generated question and CP of question in stored transformation rule,
\item pattern application rate (how many times was the pattern used),
\item pattern application percentage success (average correctness of generated patterns).
\end{itemize}
Only two components of score are permanently stored in the list (pattern application rate and pattern application percentage success). If the transformation rule is obtained from training data, these values are set to 1. If the transformation rule is derived, so we do not know the correctness of a generated question, it is based on similarity between sentences and questions. If there is other feedback (from the user), the reward is recalculated based on this feedback.
Another two components (similarity between sentences and questions) are calculated in real-time and they are not stored in the list of patterns as they are different for each input entry. The similarity calculation is shown in  table  \ref{table2}.

After the model stores the initial transformation rules learned by sentence-question pairs, it is continuously re-trained by feedback information obtained from the user. We created an online web interface where users could evaluate generated questions. These ratings help choose more reliable transformation rules for sentences. This process is also performed during the question generation phase, so it ensures the continuous improvement of the model.

Of course, there are some limitations in using transformation rules/patterns, but we adopted techniques which overcome these shortcomings. In the case of complex sentences coverage, we used a sentence simplification module which extracts simplified statements from original complex sentences as can be seen in figure \ref{figure_fw_preprocessing}. Here, a situation may occur where some extracted sentences will not be grammatically correct. However, the question generation module usually cannot generate a question from an invalid sentence as it is unlikely that there will be a pattern for it.  And, if it does create such a sentence, the generated question  usually receives a low score from the obtained by question estimation module.
The problem of low coverage in traditional pattern matching approaches is also eliminated by the distance metric, which is responsible for selecting the pattern. We do not perform exact matches in this part as usual, but instead adopt a matching process based on similarity between patterns. As we show in table 3, we can calculate the distance between two sentences (and distance between the composite pattern of a new sentence and composite patterns of all stored rules). If there is no similar transformation rule, we will obtain several rules from the hierarchy. The patterns are stored in hierarchical structure, so we can search for more abstract patterns and try to use its child patterns in this pattern hierarchy.

\subsection{Question Generation}
Once the model is created, we can use its stored set of transformation rules to generate new questions for new input sentences. Firstly, we find the set of transformation patterns applicable for new declarative sentences. They are chosen by similarity calculation between the composite pattern of a new sentence and the composite pattern of the stored transformation rule in the database. Similarity calculation reflects the difference between individual features: labels of tokens and their collocations. It can be imagined as a distance measure between two strings or a feature match in classification (calculation is shown in table \ref{table2}). Score represents the similarity between two sentences based on all patterns. There are many possibilities as to how to extend this similarity matching algorithm; we propose the basic one.

\begin{table}
  \caption{Example of Data Structure for Composite Pattern (CP)}
  \begin{minipage}{\textwidth}
    \begin{tabular}{lccccccc}
    \hline\hline
	\bf{Sent. 1:}&
	\bf{The}&
	\bf{president}&
	\bf{of}&
	\bf{Slovakia}& 
	\bf{is}&
	\bf{A. Kiska}.&
	\em{match} \\
	\bf{Sent. 2:}&
	\bf{The}&
	\bf{capital}&
	\bf{of}&
	\bf{Czechia}& 
	\bf{is}&
	\bf{Prague}. &
	 \\
        \hline
    \bf{Lemma} 				& The	& - 		& of 		& - 		& is		&-	&3/6\\
    \bf{POS} 				& DT 	& NNP	&IN 		&NNP	&VBZ		&NNP&6/6\\
    \bf{POS sim.} 			& DT		& NN 	&IN 		&NN 		&VB 		&NN 	&6/6\\
    \multirow{2}{*}{\bf{NER}}	& \multirow{2}{*}{-} 		&\multirow{2}{*}{-} 	&\multirow{2}{*}{-} 	&location/ 		&\multirow{2}{*}{-} 		&person/&\multirow{2}{*}{0/2}	\\
    		&  		& 	& 	&? 		&  		&location&	\\
    		
    \bf{SST}		& - 		&role/place	&- 	&country 		& - &-/city&1/3\\
    \bf{GKG} 		& - 		&- 	&- 	&country 		& - 		&person/city&1/2\\
    \hline
    \em{match}		&3/3 	&2/3	&3/3	&4/5	&3/3 &2/5 &\bf{34/47}	\\
    \hline\hline \\
    \end{tabular}
    \vspace{-2\baselineskip}
  \end{minipage}
  \label{table2}
\end{table}
Match score helps estimate the probability of correct usage of the transformation rule. In general, it helps us to predict the correctness of the generated questions. The algorithm of question generation procedure is shown below. There is one procedure which is not mentioned here: getTransformationPatterns. This calculates the similarity score as we described, and returns a list of transformation patterns with predicted score.

\begin{algorithm} 
\floatname{algorithm}{procedure}
\renewcommand{\algorithmicrequire}{\textbf{Input:}}
\renewcommand{\algorithmicensure}{\textbf{Output:}}
\vspace{0.1cm}
\caption{generateQuestions (sentences)}
\label{alg3} 
\begin{algorithmic}[1]
    \REQUIRE sentences
    \ENSURE questions
\vspace{0.1cm}
    \STATE $questions \Leftarrow \O$
    \FORALL{sentence \textbf{in} sentences}
    	\STATE{$cp\_sentence = createCP(sentence)$}
    	\STATE{$TP = getTransformationPatterns(cp\_sentence)$}
	\IF{$TP$} 
	    \FORALL{(tp,~score) \textbf{in} TP}
		    \STATE $questions = questions + \{(tp, score)\}$
	    \ENDFOR    
	\ENDIF
    \ENDFOR    
    \RETURN questions
\end{algorithmic}
\end{algorithm}

Questions generated by transformation rules also keep information about transformation pattern (tp) and matching score (procedure \ref{alg3}). It is used to sort and filter the best question from a group of similar questions. Questions with higher scores are considered the best ones.

\subsection{Machine Learning Approaches Used in Our Question Generation Framework}

From the perspective of machine learning, we do not use any well-known algorithm. We instead adopted an approach known as Multi-label classification and created our own distance metric for assigning new sentences into classes based on the similarity between their composite patterns. In our approach, each transformation rule can be understood as a class. Then, we can create questions for all the instances of individual classes (sentences). Sentences from the input text should be assigned to several classes and each class can create several questions. If the sentence belongs to two classes and each class generates three types of questions, then six questions will be generated and the question estimating module will sort these created questions based on its estimated quality.

 Our contribution also includes a proposed metric for distance calculation between a pair of patterns which helps us to identify possible transformation rules for input sentence. This calculation is based on the similarity between composite pattern of the new sentence and composite patterns of sentences which were assigned to this class in the training phase. 

For example, if there is a sentence with a specific structure of tokens satisfying some transformation rule (e.g. it has some entity or a specific sequence of required word types), it will be assigned to this class and the questions based on this class will then be constructed. Distance calculation for sentence classification is shown in table 3. This approach also solves the problem of low coverage in traditional pattern matching approaches, as transformation patterns are selected on the basis of general similarity. If we do not find any similar patterns, we can still find some more abstract patterns that are in the hierarchy (e.g. POS simple is just an abstracted version of the POS pattern).

In order to achieve continuous improvement of our system, we have also used some ideas from reinforcement learning approach along  with user evaluation. There are two useful concepts: solving the problem by maximizing reward function and revealing the sequence of actions which leads to final result. In our situation, we use similarity calculation as a reward function and transformation rules as a sequence of actions. 
The initial set of training data consists of sentence-question pairs which are considered to be correct, so transformation rules obtained from training data has an initial score value of 1 (maximum reward). This value is updated during the question generation phase: if an incorrect question is generated by the pattern, the total score will decrease. It is not necessary to obtain user evaluation for this. We can also use similarity calculation between new questions and already evaluated questions. In this case, the reward score for a derived transformation pattern is calculated as similarity between composite patterns of question and the score of the pattern from which the question was created. For example, if there is a pattern with a score of 0.9 and similarity between the new question and the question generated by pattern is 0.8, the final score for the new question is 0.72 (a multiplication of these values, as it shown in the equation below).

  \begin{equation}
    reward =
    \begin{cases}
      1, & \text{if}\ question\  is\  in\  training\  dataset \\
      sim(q1, q2) * score (pattern), & \text{for\ new\ question}
    \end{cases}
  \end{equation}

When we do not have user confirmation of acceptance of the question, the reward calculation in determining the suitability of applying a particular pattern is useful. Otherwise, we prefer user feedback. User interaction was conducted via simple web interface for question evaluation. Users determined if a question was acceptable (+1), unacceptable (+0) or almost acceptable (+0.5 point). Each rating was saved to the pattern. So, we keep all information about an already created question (i.e. whether it was acceptable by specific pattern). In the question generation phase, all generated questions are sorted by the question estimation module which takes this value into account. If any transformation rule obtains several zero rewards, its average score will significantly decrease in comparison with others, it will be used less times in the future and other transformation rules will be preferred. Rules with a small reward will be used only in cases when no better rule is available. However, we do not remove them completely, because information that they were not applicable to in the past may still be useful in later cases. There is also an option to utilize feedback information after real deployment of our system, so the question estimation module is constantly improved by new feedback.

\subsection{A Step-by-Step Example of Question Generation}
Here, we will demonstrate our approach on several sentences in detail. Let us suppose that we have some sentences in our training dataset:
\begin{itemize}
	\item The president of Slovakia is Andrej Kiska.
	\item The capital of Czechia is Prague.
	\item  Peter Sagan comes from Slovakia.
\end{itemize}
In the first step, users are asked to create questions which are related to these sentences. These questions should be like that:
\begin{itemize}
	\item Who is the president of Slovakia? Answer: Andrej Kiska
	\item What is the capital of Czechia? Answer: Prague
	\item Where is Prague? Answer: in Czechia
	\item Where does Peter Sagan come from? Answer: from Slovakia
\end{itemize}
After our system obtained training pairs of sentences and questions, it created composite patterns for each tuple and stored transformation patterns which keep information about the steps of transforming a sentence into question. These transformation steps are related to specific tokens of sentence, e.g.: which tokens are in a position of answer, which tokens must be moved to another position, which tokens must be removed/inserted, or which tokens must occur in an alternative form (e.g. transformation from past tense to present). We consider all information about sentence tokens (information from all patterns). As we can see in the example below, transformation from the sentence ``The president of Slovakia is Andrej Kiska.'' to the question ``Who is the president of Slovakia?'' consists of several (automatically obtained) steps: 
\begin{itemize}
	\item a sequence of tokens marked as nouns (at POS pattern) which are also marked as a Person (at NER pattern) and which occurs at the end of the sentence after the verb ``is'' will be an answer, so it will be not present in the generated question
	\item the type of a target question will be ``who'', as it refers to a question about a person, so the token will be inserted token Who will be inserted at the beginning of the question
	\item the rest of question string (after the token Who) will consist of tokens which are before the verb is
	\item etc.
\end{itemize}

All these transformation steps are obtained automatically from a sentence-question pair. This is a significant improvement over traditional pattern matching approaches. The problem with low coverage is significantly reduced as we will now explain. After finishing the training phase, our system keeps a set of transformation patterns obtained by users. When the new unseen sentence appears, it firstly tries to find several similar patterns. Of course, it is unlikely that the structure of the new sentence will be the same. The system calculates the difference between the new sentence and sentence patterns stored in the training phase and then choses several transformation patterns which are the most similar. For example, the new sentence looks like this:

\begin{itemize}
	\item Bhumibol Adulyadej was the king of Thailand.
\end{itemize}

We firstly need to compare the new sentence with sentences learned in the training phase. Let’s look at the comparisons:

Comparison with the composite pattern of sentence 1: \textit{The president of Slovakia is Andrej Kiska.}

\begin{itemize}
	\item there is also a token sequence which represents person, but in a different position
	\item there is also a token which represents location, but in a different position
	\item the same verb is used, but in a different tense
	\item there is the same preposition before the token of location
	\item there is a noun with a similar meaning before the location (both tokens represent ``role'' at semantic level: king and president)
\end{itemize}
Comparison with the composite pattern of sentence 2: \textit{The capital of Czechia is Prague.}
\begin{itemize}
	\item there are two tokens which represent location, one of which is in the same position
	\item there is also a preposition ``of'' before a token of location
	\item there is also a noun before the sequence of ``of location''
	\item there is no token of person (so it will be impossible to generate a question of type ``Who'')
\end{itemize}
Comparison with the composite pattern of sentence 3: \textit{Peter Sagan comes from Slovakia.}
\begin{itemize}
	\item there is also a token sequence which represents person in the same position
	\item there is also a token which represents location in the same position
	\item there is also a preposition between token of location, and it is similar (from vs. of)
	\item there is a verb before preposition and location
\end{itemize}

As we can see from the comparisons, the new sentence (Bhumibol Adulyadej was the king of Thailand) is most closely to the first sentence (The president of Slovakia is Andrej Kiska). Despite the positions of location tokens and person tokens being at opposite ends of the sentences and the verbs being in different tenses, both sentences can be covered by the same transformation rule as we show on example below:

\begin{itemize}
	\item sentence from training dataset: The president of Slovakia is Andrej Kiska.  
	\item question: Who is the president of Slovakia? Andrej Kiska.
\end{itemize}

\begin{itemize}
	\item sentence from testing dataset: Bhumibol Adulyadej was the king of Thailand.  
	\item question: Who was the king of Thailand? Bhumibol Adulyadej.
\end{itemize}

In this way, our AQG system firstly finds the most similar patterns and then generates all possible questions, then evaluates them by an estimation module and finally chooses several of them based on highest estimated quality.

%
%
%
%

\section{Experiments and Evaluations}
In order to provide evaluation of our approach, we needed datasets for training and testing. As stated previously, the widely-used dataset QGSTEC is too small to be manually divided into training and testing parts, and it is also of insufficient size for training. We therefore decided to use it only for the evaluation of generated questions and for the mutual comparison of AQG systems. For the training phase, we created our own dataset consisting of sentence-question pairs extracted from Wikipedia articles. Questions were obtained from users: we created a user interface and users could assign questions to randomly selected articles shown therein. We obtained about 1,200 pairs of sentences and questions as an initial set for training. We then used our trained model on the QGSTEC dataset and compared generated questions with questions generated by state-of-the-art systems. We also imported questions generated from published systems, made a deeper comparison of questions, and tried to calculate the difference between questions generated by AQG systems with questions expected by evaluators.

\subsection{Our First Experiments}
We will briefly describe our first experiments. Whereas it was our first attempt to compare questions across the systems based on our previous research \cite{blstak2017}. We consider the later experiments (described in the next section) as more valuable. In these first experiments, we took evaluation data from participants of the question generation task challenge. Questions generated by participants were evaluated from various points of view (correctness, ambiguity, variety) by human evaluators on a four-step scale. We converted correctness values to a percentage scale and the number of correct questions was calculated by average correctness of questions and all generated questions. This was done because it is difficult to assign correctness value on a large scale while keeping the same evaluation criteria (questions from our system were divided into two categories: correct and incorrect). If there were some groups of similar questions, only one question was included in the results. This is also the reason why we generate less questions in comparison with participants of QGSTEC, but as shown in table \ref{table3}, the average correctness of our questions was much better.

\begin{table}
  \caption{Comparison of AQG approaches on QGSTEC dataset}
  \begin{minipage}{\textwidth}
    \begin{tabular}{lcccc}
    \hline\hline
	System &
	Correctness (\%) &
	Quest. / sentence &
	Correct q. &
	Generated q. \\
    \hline
    Participant 1\footnote{\cite{yao2010}} 		&0\hpt64 &4\hpt37 &227 &354 \\
    Participant 2\footnote{\cite{varga2010}} 		&0\hpt34 &2\hpt04 &56 &165 \\
    Participant 3\footnote{\cite{pal2010}} 		&0\hpt30 &2\hpt58 &63 &209 \\
    Participant 4\footnote{(Ali et al. 2010)} 		&0\hpt21 &2\hpt07 &35 &168 \\
    Curto et al\hpt\footnote{(Curto et al. 2012)} 	&0\hpt56 &0\hpt31 &14 &25 \\
    Our approach\footnote{\cite{blstak2017}} 	&0\hpt80 &1\hpt50 &98 &121 \\
    \hline\hline \\
    \end{tabular}
    \vspace{-2\baselineskip}
  \end{minipage}
  \label{table3}
\end{table}

As we can see, participants of the question generation task challenge generated more questions, but average correctness was significantly lower. The larger number of questions generated by QGSTEC participants can also be justified by the fact that one of the evaluation criteria was the total number of generated questions and participants had information about the maximum number of possible generated questions. In case of participant 1, many duplicates or very similar questions were generated. Only 320 questions (approximately 90 per cent) were left after removing the duplicates. The drawback of Curto’s system is that it was trained on data about artefacts in a museum only, so the coverage of other topics was weak.

\subsection{Comparison of Generated Questions}
We realized other experiments for better comparison of AQG systems. Fortunately, generated questions from QGSTEC systems are available for download, so we created our own interface for question evaluation, imported all questions to it and our users evaluated all questions. So, the comparison conditions were the same for all obtained results. We also published our reporting interface where we can see all results and data obtained from evaluators (from all experiments mentioned in this work). It can be seen on our site in the section \emph{Evaluations} \footnote{available at \url{http://www2.fiit.stuba.sk/~blstak/aqg/index.html} }. 

Our users evaluated questions generated by our system and questions generated by the two best AQG systems of QGSTEC event. Questions (with source sentences) were randomly shown on a simple interface and users could evaluate two properties: grammatical correctness (syntax) and semantical correctness (interpretability, clarity) on a three-step scale (correct, partially correct and incorrect). There was also an option not to respond (skipping the question) as some questions required a good knowledge of English. If the question was not marked as correct, a user could write correction (if she/he wants). From our point of view, this type of evaluation is significantly easier for users. 

There were a total of 706 questions in our interface for evaluation. We carefully analyzed questions generated from the testing part of dataset (585 questions). The remaining 121 questions were generated by our system on training dataset and they were there for two purposes: (1) to increase the variety of questions and to suppress user fatigue from evaluating a large amount of similar questions and (2) to compare training and testing datasets. We could not use the training dataset for the training phase due to its small volume (from the nature of event it is obvious, that the training dataset served only as sample data for participants). In total, we obtained 2,739 ratings and the results are shown in table \ref{table4}.

\begin{table}
  \caption{Comparison of AQG approaches on QGSTEC dataset by user evaluators in our interface}
  \begin{minipage}{\textwidth}
    \begin{tabular}{lccccccc}
    \hline\hline
	system&
	questions& 
	evaluat-& 
	avg.\footnote{average score}&
	syn.\footnote{average syntactic score}&
	sem.\footnote{average semantic score}&
	correct &
	correctness \\
	& &ions & score & score & score & quest. & (\%)
	\\
     \hline
    A 		&320 &1216&0\hpt19 &0\hpt21&0\hpt17&453&37 \\
    B 		&165 &641 &0\hpt47 &0\hpt49&0\hpt45&335&52 \\
    M 		&100 &403 &0\hpt71 &0\hpt74&0\hpt68&277&69 \\
    M train 	&121 &479 &0\hpt73 &0\hpt75&0\hpt70&322&67 \\
    \hline
    Sum 	&706 &2739 &  & & &1387 \\
        \hline\hline \\
    \end{tabular}
    \vspace{-2\baselineskip}
  \end{minipage}
  \label{table4}
\end{table}
\begin{figure}
  	\vspace{0.4cm}
	\caption{Histogram of votes (syntactic + semantic) obtained by user evaluators for each AQG system: our system (marked as M) obtained more positive points and less negative points in comparison to other state-of-the-art systems.}
	\includegraphics[totalheight=8.2cm]{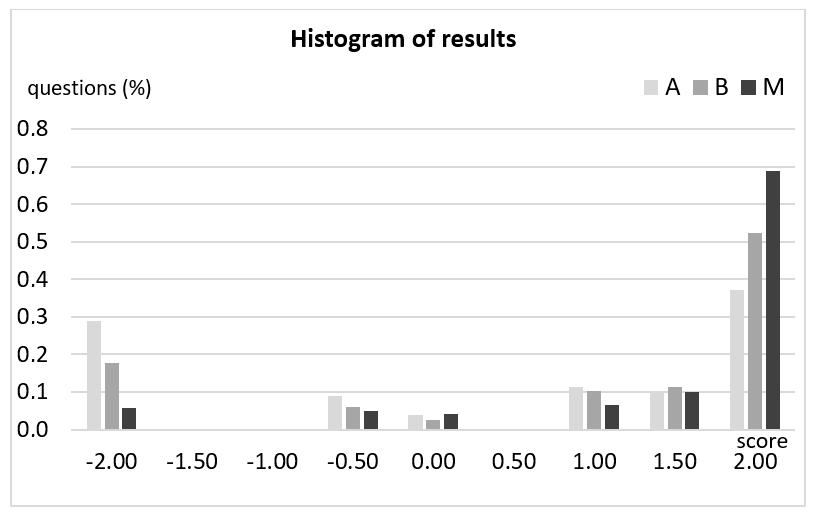}
	\label{figure2}
\end{figure}

Our system (marked as M) generates more correct questions and less incorrect questions in comparison to the other systems (marked as A and B which correspond to the first two systems from our previous experiment). On the other hand, our system generates a smaller amount of questions. The larger amount of generated questions by participant A is caused by the fact that for each sentence it generates four questions regardless of whether or not it was possible. The total correctness of questions generated by systems of other participants would be even lower in real text, because if we randomly choose any sentence from the article, it is not common to create four questions from this sentence. On the contrary, there are many sentences for which no question can be created.

We also evaluated the ability to cover various types of sentences in these systems. Compared systems were often forced to generate a specific number of questions regardless of quality. Our coverage metrics show the number of sentences assigned to at least one question with a certain value of correctness. 

\begin{table}
  \caption{Coverage of sentences with generated questions based on question correctness}
  \begin{minipage}{\textwidth}
    \begin{tabular}{lccccccc}
    \hline\hline
	system coverage&
	2& 
	1.5+& 
	1.0+& 
	0.0+& 
	all && \\
     \hline
    A 		&1\hpt11 &17\hpt78&27\hpt78 &67\hpt78&98\hpt89 \\
    B 		&14\hpt44 &33\hpt33&44\hpt44 &67\hpt78&84\hpt44 \\
    M 		&17\hpt78 &37\hpt78&47\hpt78 &50\hpt00&93\hpt33 \\
        \hline\hline \\
    \end{tabular}
    \vspace{-2\baselineskip}
  \end{minipage}
  \label{table5}
\end{table}

As we can see in table \ref{table5}, coverage of our system (M) is better when we focus on question quality (maximum value for correctness was two points: one point for syntactic correctness and one point for semantic correctness). 

Whereas there were more user evaluators, and everybody has their own opinion about correctness, it is also interesting to compare their evaluations against each other. We calculated Inter-rater reliability (IRR), which is the metric used for qualitative data analysis to obtain the degree of agreement among raters. Each question obtained several ratings by various users. There were ten evaluators and the minimum number of votes for question was three, with the maximum number of votes being five. Although it was on a numeric scale, we did not used a standard version (which uses Cohen`s Kappa), but rather two modified versions:  
\begin{itemize}
 \item IRRb (binary IRR) – reflects the number of consistent evaluations (same vs. different) 
 \item IRRn (numeric IRR) – reflects the numeric difference between all evaluations 
\end{itemize}
We know, that percentage IRR overestimates the level of agreement and it does not take chance agreement into account, but we used the results only for comparisons, not for qualitative data analysis of the dataset. The results are shown in table \ref{table6}: independently for syntactic votes, semantic votes and then their average obtained score. 

\begin{table}
  \caption{Inter-rater reliability of evaluators}
  \begin{minipage}{\textwidth}
    \begin{tabular}{lcccccccc}
    \hline\hline
	system& &
	IRRb&
	IRRb&
	IRRb& &
	IRRn & 
	IRRn &
	IRRn \\
	& &
	syn&
	sem&
	average& &
	syn&
	sem&
	average
	\\
     \hline
	A& 	&	62\hpt60 &61\hpt55&62\hpt08& &76\hpt79&77\hpt14&76\hpt96 \\
	B& 	&	62\hpt95 &62\hpt98&62\hpt96& &78\hpt03&77\hpt57&77\hpt80 \\
	M& 	&	62\hpt24 &71\hpt30&69\hpt27& &82\hpt67&85\hpt93&84\hpt30 \\
	average&& 	63\hpt49 &63\hpt62&63\hpt56& &78\hpt15&78\hpt76&78\hpt46 \\
	\hline\hline \\
    \end{tabular}
    \vspace{-2\baselineskip}
  \end{minipage}
  \label{table6}
\end{table}
From this experiment we can see, that evaluators have agreed in about 63 per cent (IRRb) and 78 per cent (IRRn). In both cases, the total agreement was better during evaluation of questions using our system.

\subsection{Acceptance of Generated Questions}
Finally, we made a comparison between questions generated by AQG systems and questions expected by users (marked as reference questions). During the previous evaluations, users had the option to write a question appropriate for them.  We then compared these tuples using metrics used for sentence comparisons in related NLP tasks: BLEU - BilinguaL Evaluation Understudy and ROUGE - Recall-Oriented Understudy for Gisting Evaluation. There are several metrics in the area of Natural Language Generation which are used for the evaluation of generated text \cite{gatt2017} and we chose BLEU and ROUGE as they are metrics based on phrase-matching and also very popular in similar tasks such as machine translation and text summarization.

The purpose of this experiment was to reveal how different the questions generated by AQG systems were in comparison to questions created by users. We firstly filtered all questions generated by the systems with a score greater than 75 per cent and calculated the difference between these and questions created by users. If there were more questions assigned to one sentence, we chose the most similar one to decrease the total difference. Example of similarity metrics between two sentences, where the difference lies in two words are shown below:

\vspace{6px}
\begin{tabular}{lp{6cm}}
\multicolumn{2}{l}{\em Similarity calculation for sentence}       \\
\\
	generated question:& 	Is Egypt situated in the north? \\
	reference question:& 	Is Egypt situated in the north of Africa? \\
	BLEU - average:&		0.72 \\
	ROUGE-L:&			0.84 \\
\end{tabular}
\\

\begin{table}
  \caption{Comparison of reference questions (expected by users) and generated questions by AQG systems}
    \begin{tabular}{lcccccccc}
    \hline\hline
	system&
	A&
	B&
	M &&&&&  \\
	number of questions & 	122 & 89  & 66  \\
     \hline
	BLEU - 1gram 	& 	0\hpt76 & 0\hpt81 & 0\hpt79  \\ 
	BLEU - 2gram 	& 	0\hpt71 & 0\hpt76 & 0\hpt75  \\
	BLEU - 3gram 	& 	0\hpt66 & 0\hpt72 & 0\hpt72  \\
	BLEU - 4gram 	& 	0\hpt61 & 0\hpt69 & 0\hpt70  \\
	BLEU - average  & 	0\hpt69 & 0\hpt75 & 0\hpt74 \\
	ROUGE-L	 	& 	0\hpt77 & 0\hpt81 & 0\hpt86 \\
      \hline 
	length - reference questions 	& 5993&	5456&	2694\\
	length - generated questions 	& 6278&	5903&	2382\\
	length ratio				& 1\hpt06&	1\hpt08&	0\hpt87\\
      \hline\hline  \\
    \end{tabular}
    \vspace{-2\baselineskip}
   \label{table7}
\end{table}
There are no significant differences among the evaluated systems, but there are some significant aspects resulting from these evaluations. Our system generates slightly better questions based on BLEU and ROUGE-L metrics, which means that questions generated by our system are more expected (they look more like questions created by a teacher). The last rows show question length (in characters). On average, the length of questions generated by our system was shorter than the length of expected questions. Other systems usually generated questions with some (possibly) redundant words. 

\subsection{Evaluations on SQuAD dataset}

The most recent papers use SQuAD dataset for question evaluation. As we mentioned before, this dataset is originally created for the opposite task (Question Answering Dataset) and this presented several limitations for us. We evaluated our system on this dataset as it was prepared for the question generation task in Du et al. \shortcite{du2017}\footnote{\url{https://github.com/xinyadu/nqg}}. We imported articles from the test set to our system and made preprocessing steps as described in section 3.1. Our system generated about 830 questions and our question estimating module selected 100 questions based on highest estimated quality. We then evaluated the quality of the generated questions with human assessors (table \ref{tablecomparisonsquad1}). It is difficult to compare human scores from different papers due to different evaluation criteria. We therefore kept criteria from our previous evaluation on QGSTEC, which are also totally consistent with Kumar et al.\shortcite{kumar2018}. In other works the criteria are marked differently. The average score (the last column in the table) is calculated based on all metrics (penultimate column).

\begin{table}
  \caption{User evaluation on SQuAD dataset}
  \begin{minipage}{\textwidth}
    \begin{tabular}{lcclcc}
    \hline\hline
	System &
	Questions &
	Scale &
	Score values &
	Average \\
    \hline
    Du et al. \shortcite{du2017} 				&100 	&1--5 &78\hpt2\% (naturalness: 3\hpt91/5)  &65\hpt4\% \\
		& & & 52\hpt6\% (difficulty: 2\hpt63/5) \\
    \hline

    Zhou et al. \shortcite{zhou2018}				&200 	&1--3 &72\hpt6\% (score: 2\hpt18/3)	&72\hpt6\% \\
    \hline

    Kumar et al. \shortcite{kumar2018} 			&100 	&0--100 &84.0\% (syntax) &81\hpt2\% \\
		& & &81.3\% (semantics) \\
		& & &78.3\% (relevance) \\
    \hline
    Hosking et al. \shortcite{hosking2019} 		&300 	&1--5 &66\hpt8\% (fluency: 3\hpt34/5) &64\hpt6\% \\
												& & &62\hpt4\% (relevance: 3\hpt12/5) \\
    \hline

    our system  								&100 	&0--100 &86\hpt5\% (syntax) &81\hpt6\% \\
		& & &76.8\% (semantics) \\
    \hline\hline \\
    \end{tabular}
    \vspace{-2\baselineskip}
  \end{minipage}
  \label{tablecomparisonsquad1}
\end{table}

We then made evaluation leveraging BLEU and ROUGE metrics as previously. However, it is not possible to make a direct comparison of our approach with neural network models. This is because our system uses a question estimator module which removes questions with low score of acceptance. It gives priority to questions which are estimated as better even if they are different to the target question in the dataset. Our system does not generate questions which are expected but unreachable. Neural network models are based on the idea of generating questions which are as similar as possible to the target questions regardless of correctness. They just try to fit question structure as close as possible to that in the dataset even if the question is not correct. As we can see from the evaluation table in the state-of-the-art section \ref{tablecomparison2}, these questions are far from correct. Unfortunately, there are many questions in this dataset which cannot be generated without external knowledge because  interconnection to the fact is missing. As an example, the question-sentence pair in the dataset looks like this:
\begin{itemize}
\item Question: What is in front of the Notre Dame Main Building?
\item Sentence: Immediately in front of the main building and facing it, is a copper statue of Christ.
\end{itemize}
There is also a full paragraph linked to this question-sentence pair in the dataset as context information, however the token Notre Dame is not mentioned anywhere. Without external knowledge that this question is about Notre Dame, it is not possible to construct this question. In our system, this type of question is filtered out whereas a question estimator module can determine that they are not correct when some information is missing. On the other hand, our system generates questions which are correct but not present in the dataset. So, our evaluation is done with the intent of evaluating questions from the view of correctness, not whether they are present in the dataset as target questions.

\begin{table}
  \caption{Comparison of reference questions (expected by users) and generated questions by our AQG system on SQuAD dataset}
    \begin{tabular}{lc}
    \hline\hline
	BLEU - 1gram 	& 	0\hpt83 \\ 
	BLEU - 2gram 	& 	0\hpt78 \\
	BLEU - 3gram 	& 	0\hpt74 \\
	BLEU - 4gram 	& 	0\hpt71 \\
	ROUGE-L	 	& 	0\hpt87 \\
      \hline 
	length - reference questions 	& 3553\\
	length - generated questions 	& 3274\\
	length ratio				& 1\hpt08\\
      \hline\hline  \\
    \end{tabular}
    \vspace{-2\baselineskip}
   \label{tablecomparisonsquad2}
\end{table}


As we can see in table \ref{tablecomparisonsquad2}, BLEU and ROUGE values on the SQuAD dataset are similar to the QGSTEC dataset (table \ref{table7}) in the previous section. Our system is able to generate correct questions without any modification (we have used the same settings as we described earlier). However, a direct comparison of the results with other systems in this case would not be correct as the generated questions are not the same.

\subsection{Comparison of questions based on usefulness}

The questions generated by the patterns differ from the questions generated by the neural networks (NN) in several aspects. Although it is difficult to say which questions are globally better, we will try to highlight the differences revealed by our analysis and experiments. We have compared the questions generated from the Squad dataset, and since the comparison also required manually obtained user ratings, we selected questions from the first few sections of the test partition of Squad dataset. In case of questions generated by neural networks, we used exports from Du et al. \shortcite{du2017} available online on Github as in evaluation in the previous section.

We have adopted modified evaluation scale used in \shortcite{skalban2013} and let the evaluators classify the generated questions based on their usefulness into following categories:
\begin{itemize}
\item 3: question may be used without change
\item 2: question may be used with small correction
\item 1: question is useful for inspiration 
\item 0: not sure
\item -1: question is not useful
\end{itemize}
Based on this comparison we can summarize the differences between questions generated by different approaches, but it is difficult to calculate their suitability on one scale. The traits of our approach in comparison to NN approaches are:
\begin{enumerate}
\item our system generates lower number of questions, but significantly less duplicates and similar questions,
\item our system generates more simple questions (without useless information), 
\item our system generates more questions which can be used without change (after ordering the question based on estimated value)
\end{enumerate}
The first statement was calculated as a number of questions marked as duplicated by evaluators. There were about fifteen per cent of questions marked as unusable due to its similarity with some previous questions. The second statement was calculated by the number of tokens in sentences. The average difference was about 2.5 tokens. Results are obtained from the experiment in table 11: the last three rows show the difference in sentence length in comparison to an optimal question. And the third statement: our system generated about 50 per cent of questions usable without modification on Squad dataset (48 questions from 100). This is also obtained from the experiment shown in table 11. As we have mentioned before, no questions generated by NN were usable without change (without post-processing modification).

However, numeric evaluation has failed due to several reasons and therefore we do not report the results in details. From our point of view, it is difficult to ensure fair rules for the comparison of the questions generated by different approaches. We will describe the main reasons in the following paragraphs.
Firstly, there were no absolutely correct questions generated by NN (with score 3). It is because there are little grammar issues in each question generated by NN (e.g. incorrect capital letter, lowercase letters used in acronyms, spaces before the question marks and commas etc.). It is caused by common pre-processing steps. Of course, these little grammar issues can be fixed by post-processing steps. However, this would affect evaluation and it is difficult to determine which grammatical inaccuracy should still be tolerable. On the other hand, many of our questions (generated by transformation rules) are absolutely correct due to patterns which also do post-processing correction automatically. In NN approaches dealing with text, it is typical to train the word embeddings and character level embeddings models on the alphabet without uppercase letters. Evaluation by standard metrics (e.g. BLEU) has minimal impact on this type of evaluation.
Another problem  is associated with duplicates and very similar questions. As NN models generate more questions, they also generate more similar questions. It is not easy to create instructions for the evaluators, which would guide them how to take into account duplication and if it should be skipped or not. Our system has a separate module which eliminates duplicates and similar questions, so the total number of generated questions is lower. 
There are also differences in structure of generated questions. Questions generated by neural networks are usually longer with more words. This type of question is usable, however sometimes it keeps irrelevant information. Here are some examples from article about Notre Dame (output from NN model without modification):
\begin{itemize}
\item in 1919 a new president of notre dame was named , who was it ?
\item on what date was the rebuilding of the main building begun at notre dame after the fire that claimed the previous ?
\item over how many years did the change to national standards undertaken at notre dame in the early twentieth century take place ?
\item around the time that rev. cavanaugh became president of notre dame by how much did the undergrad student body of notre dame increase ?
\end{itemize}
From the point of view of evaluation, users had trouble deciding whether to mark these questions as unusable or as suitable for inspiration, as they contain several facts to create some questions. Our system penalized questions that did not match the grammatical structure of the standard question (as we described in the question estimation module of our system in section 3.2).
Several advantages of our system could probably be applied also to the questions generated by NN, but it would require the additional post-processing steps (e.g.: correction of letters at the beginning of the question, remove useless spaces, checking the presence of the interrogation pronoun, checking the length of the sentence, similarity calculation of the syntactic structure of the question with the structure of a typical question etc.).

\subsection{Generated questions in real-world scenarios}
In this section, we focused on evaluating the applicability of the questions in real world scenarios: how useful the automatically generated questions are. Since the main goal of the questions in the educational process is to verify the students' knowledge, we have decided to evaluate the suitability of the questions considering two points of view:

\begin{itemize}
\item How useful are generated questions for the lecturers? (How the generated questions help lecturers to create a test.)
\item How useful are generated questions for the students? (How the generated questions help students to verify their knowledge.)
\end{itemize}
For these experiments, we used the same sets of questions as in experiments before. Students had to determine if they thought that the questions were appropriate for verifying their knowledge. On the other hand, lecturers had to determine if the proposed questions are suitable for using in testing. In both cases, we chose a four-level scale:
\begin{itemize}
\item The question is not usable for this purpose (0 points).
\item The question can serve as a guide for creating new suitable questions (1 point). 
\item The question is suitable for this purpose with little correction (2 points).
\item The question is suitable for this purpose without change (3 points).
\end{itemize}
The results of this experiment are shown in table \ref{questionusable}. In the first column (marked as average points), there is a rounded value of average score. In the last row, there is an average score for the full dataset. Experiment was done on the dataset of 98 questions (there were originally 100 questions, but two of them turned out to be too similar, so we decided to remove them from numerical evidence). The results show that about 45 questions were usable in real application. There were no significant differences across groups, students and lecturers indicated the usability of questions relatively similarly. The average points (shown in the last row) are calculated as a relative score across all groups.

\begin{table}
  \caption{Evaluation in real scenarios}
  \begin{minipage}{\textwidth}
    \begin{tabular}{ccccccc}
    \hline\hline
	Average points  &  
	students & students &
	lecturers & lecturers &
	both & both  
	\\
	(rounded) &  
	total &  
	percent &
	total &
	percent &
	total &
	percent  
	\\
    \hline
    \hline
    
    3 & 45 & 45\hpt9\% & 47& 47\hpt8\% & 44 & 44\hpt9\% \\    \hline
    2 & 44 & 44\hpt9\% & 46& 46\hpt9\% & 46 & 46\hpt9\% \\    \hline
    1 &  8 &  8\hpt2\%  &  4	&  4\hpt1\%  &  6 & 6\hpt1\% \\    \hline
    0 &  1 &  1\hpt0\%  &  1 &  1\hpt0\%  & 2 & 2\hpt0\% \\
    \hline
    \hline
    Average & 2\hpt33 & 77\hpt7\% & 2.38 & 79\hpt3\% & 2\hpt35 &  78\hpt2\% \\
    \hline\hline \\
    \end{tabular}
    \vspace{-2\baselineskip}
  \end{minipage}
  \label{questionusable}
\end{table}

\subsection{Evaluating the Usefulness of the Generated Questions for Self-education}

Automatically generated questions might be useful in several ways. Our main goal was to create questions for educational purposes – i.e. students could use these questions to improve their text comprehension skills or ability to memorize facts. 

To verify the benefits of generated questions, we performed an experiment aiming to prove the following statement: the questions automatically generated from the text can help to keep in mind more information from educational materials in comparison to educational materials without questions or educational materials with just some generic questions. We prepared several articles for reading from the encyclopedia Wikipedia. Then we tried to find out if the students remembered more facts and assertions introduced in this text in case that the particular questions were available below the text (these particular questions were directly related to the facts in text). We have chosen six short articles (from 85 to 210 words) about countries from our previous experiments \cite{blstak2017}. We have selected countries that are mostly unknown to the students  (e.g. Algeria or Brunei) to prevent them from already knowing the facts before reading the article. Students had to confirm that they did not know much about these countries. We also slightly changed some trivial statements to ensure that students have responded according to information in the article and not according to their general knowledge. Articles were ordered randomly and there were three groups of them for each student:

\begin{itemize}
\item In the first group of articles, there was just a heading and a text (abstract) without any questions in the user interface. 
\item In the second group of articles, there were several generic questions followed by an article (e.g. Was this article interesting to you?).
\item In the third group, the articles were enriched by several questions generated by our AQG system without modification. 
\end{itemize}
After the reading session, students were asked to write the facts which they have remembered. We also asked for confirmation that students have read available questions below articles (if there were any). The average number of remembered facts is shown in the table below (table \ref{extrinsic2}). Each row is corresponding to one of the groups of articles. It would not make sense to show the score for individual articles, because each student had a different set of articles in each group.

\begin{table}
  \caption{Extrinsic evaluation: measuring the suitability of questions in the process of self-education}
  \begin{minipage}{\textwidth}
    \begin{tabular}{lcccccccc}
    \hline\hline
	&
	\multicolumn{6}{c}{Score (\%)} 	& 
	Average (\%) \\
 	Question group &
	S1 & 
	S2 & 
	S3 & 
	S4 &  
	S5 &  
	S6 & 
	& \\
    \hline
    Group 1\footnote{articles without any question} 							& 55\hpt0 &55\hpt0 &66\hpt0 &67\hpt5 &50\hpt0  &72\hpt5  & 61\hpt0 \\
    Group 2\footnote{articles with generic questions} 							& 67\hpt5 &41\hpt5 &65\hpt0 &67\hpt5 &72\hpt5  &70\hpt0  & 64\hpt0 \\
    Group 3\footnote{articles with factual questions generated by our system} 	& 73\hpt5 &65\hpt0 &70\hpt0 &75\hpt0 &77\hpt5  &70\hpt5  & 72\hpt0 \\
    \hline\hline \\
    \end{tabular}
    \vspace{-2\baselineskip}
  \end{minipage}
  \label{extrinsic2}
\end{table}

As we expected, the participants remembered more facts from articles in which also some questions were included. We have also examined the  reading strategy of our participants. Almost every participant has read all articles more than once (usually twice). However, after they have read questions below the text, they usually paid more attention to the  statements related to these questions. Before the second reading, students usually checked whether  they knew the answers to the questions below the text. In the case of general questions (Group 2), they did not need to read anything again as the question did not cover any specific part of the text – so they could answer these questions almost arbitrarily (that was the intention). Only the questions in Group 3 were content-specific, and so it exhorted participants to subconsciously verify the answer from the given text (and sometimes to read it again). As a part of the evaluation, we also analyzed the differences across the articles' score. The longest article had the weakest success in terms of percentage coverage of facts, however this did not apply to the whole dataset. It often happened that longer articles obtained better scores. Based on discussions with the students, this may have been due to the fact that with more statements available in the article it was also easier to remember more information in total. In the case of extremely short articles, every unmentioned fact caused high point reduction in total average score.

Based on our experiment, we can confirm that the questions shown below the text are beneficial and should be used in self-educational process. Also, automatically generated questions by our system support this process. Factual questions are more suitable for knowledge memorization in comparison to just general types of questions (e.g. What was this article about?). They also  encourage students to think about the content of the text, to search patiently for the correct answer and thus to verify obtained knowledge more thoroughly.

%
%
%
%

\section{Conclusions and Future Work}

In this paper we presented our novel approach to the task of automatic question generation from unstructured text. It combines a linguistic approach based on various types of sentence patterns (lexical, syntactic and semantic patterns) with machine learning approaches. Transformation rules which transform declarative sentences into questions are automatically obtained by an initial set of sentence-question pairs. The advantage of this approach lies in the simple expansion of transformation rules and continuously improving the model. Therefore, intervention by human experts is less necessary than with other approaches.

The main contributions of our method can be summed up into following points: Firstly, a data-driven approach allows the model to train and retrain question generation processes by sentence-question pairs, so the framework does not need human experts on linguistics to create transformation rules manually or human experts on software engineering to modify the source code. Improvement of the system is based on explicit feedback from users, which is used in reinforcement learning and by adding new sentence-question pairs to the model (supervised learning). Finally, a hierarchy of patterns allows the framework to create various types of questions according to the similarity match between sentences on various levels of processing (linguistic, syntactic and semantic level).

We conducted several experiments to evaluate our method and to compare the quality of generated questions. At first, we compared the correctness of the generated questions and the total number of correct questions generated by our system with several state-of-the-art systems. We used a question generation dataset for this purpose and the results have demonstrated that our approach generates a greater number of acceptable questions. We also imported our generated questions and questions generated from the two best systems (based on evaluation of a QGSTEC event) into an online interface and let users evaluate all questions in random order. Users could evaluate the quality of questions and they could also write their own questions related to the sentences. Then we compared questions suggested by users with questions generated by question generation systems using BLEU and ROUGE metrics. Finally, we also evaluated our system on the dataset SQuAD and confirmed that the system is able to create questions for texts from this dataset, with similar correctness as before. Comparing the generated questions with other systems was not straightforward in this case, as our system generated different types of questions. Generated questions reached the required correctness, but the questions were not the same, so direct comparison is difficult. Finally, we have also analyzed differences between questions generated by neural networks and questions generated by our system in terms of usability in real world scenarios. The results show that our contribution to the task of question generation is meaningful.

We have several ideas how to improve and functionally expand our framework in the future. We are using semantic information about tokens obtained mainly from named entity databases (e.g. Viaf or Google Knowledge Graph), but it would be interesting to leverage global word taxonomy and classify nouns, adjectives or verbs into several categories. These word categories should be used as features for machine learning and pattern filtering. We also plan to leverage a related NLP task called coreference resolution and identify all occurrences of the same entity. This would be beneficial in increasing sentence coverage as many sentences contain references to subjects (by pronoun or acronym). Of course, it will be interesting to verify if our approach is applicable to languages other than English. One of the possible approaches which could bring benefit in the future is to explore the utilization of transfer learning \cite{pikuliak2021}.

\subsubsection*{Acknowledgments}
This research was supported by TAILOR, a project funded by EU Horizon 2020 research and innovation programme under grant agreement No 952215.

\appendix

\label{lastpage}

\end{document}